%% file: main.tex
\documentclass{article}
\usepackage{M-PSI}

\usepackage{tikz}
\usetikzlibrary{shapes}
\usepackage{cite}
\usepackage{amssymb,amsfonts}
\usepackage{fontawesome}
\usepackage{algorithmic}
\usepackage{graphicx}
\usepackage{textcomp}
\usepackage{xcolor}
\usepackage{url}
\usepackage{multirow}
\usepackage{tabularx,ragged2e}

\usepackage{enumitem}
\def\BibTeX{{\rm B\kern-.05em{\sc i\kern-.025em b}\kern-.08em
    T\kern-.1667em\lower.7ex\hbox{E}\kern-.125emX}}

\usepackage[switch]{lineno}

\title{Knowledge Distillation for LLM-Based Human Activity Recognition in Homes}

\author{Julien Cumin\inst{1} \and Oussama Er-Rahmany\inst{1} \and Xi Chen\inst{1,2}}

\author{\href{https://jcumin.github.io/}{Julien Cumin\textsuperscript{1~\small{\ExternalLink}}}, Oussama Er-Rahmany\textsuperscript{1}, Xi Chen\textsuperscript{1,2}\vspace{0.1cm}\\
{$^1$ Orange Research}\\
{$^2$ Univ. Grenoble Alpes, CNRS, Grenoble INP, LIG, 38000 Grenoble, France}\\ %
{Corresponding author: \href{mailto:julien1.cumin@orange.com}{julien1.cumin@orange.com}}
}

\usepackage[pdfencoding=auto, pdfusetitle]{hyperref}
\hypersetup{
  hidelinks,
  urlcolor=red,
  pdftitle={Knowledge Distillation for LLM-Based Human Activity Recognition in Homes},
  pdfauthor={Julien Cumin, Oussama Er-Rahmany, Xi Chen},
  pdfkeywords={Human activity recognition, large language models, knowledge distillation, ambient intelligence, smart homes.},
}

\begin{document}


\begin{abstract}
Human Activity Recognition (HAR) is a central problem for context-aware applications, especially for smart homes and assisted living. A few very recent studies have shown that Large Language Models (LLMs) can be used for HAR at home, reaching high performance and addressing key challenges. In this paper, we provide new experimental results regarding the use of LLMs for HAR, on two state-of-the-art datasets. More specifically, we show how recognition performance evolves depending on the size of the LLM used. Moreover, we experiment on the use of knowledge distillation techniques to fine-tune smaller LLMs with HAR reasoning examples generated by larger LLMs. We show that such fine-tuned models can perform almost as well as the largest LLMs, while having 50 times less parameters.

\end{abstract}

\section{Introduction}

Human Activity Recognition (HAR) in homes based on sensor data has been a long-standing subject of research, with potential applications for healthcare, assisted-living, home security, energy management, and smart home environments in general \cite{crowley2015ecological}. Many different kinds of sensors have been historically used for HAR: ambient sensors that measure changes in the environment (e.g. motion, door openings, etc.), cameras, microphones, wearable sensors, etc. \cite{arrotta2024multi}. We focus mostly on ambient and wearable sensors in this work. 

Developing HAR models in such environments is still challenging to this day, for a wide variety of reasons. First of all, collecting and labeling representative data of human activities in homes is very expensive and time-consuming. Moreover, each individual home vastly differs from the next, with different room layouts, sensor installations, occupant routines, family structures, etc. Those two challenges prevent the development of general-purpose HAR systems that can generalize to any possible home \cite{chen2024towards}.

In addition, multi-subject situations, where multiple occupants perform activities separately or conjointly in the home, create new challenges: sensor data is no longer attached to a singular subject and activity, but can be the result of interleaved sensor events triggered by multiple activities \cite{arrotta2023micar}. In any case, such situations must be addressed by current studies, since they are representative of most real-world home situations.

Recently, a few new approaches based on Large Language Models (LLMs) have been proposed to tackle the problem of HAR in homes \cite{civitarese2025large, chen2024towards}. LLMs already hold wide knowledge of human behavior and situations of life at home, and many recent LLMs have been pretrained to follow ``chain-of-thought'' reasoning patterns. These characteristics can be leveraged to use an LLM for HAR through prompt engineering, circumventing the difficulties of collecting training data and generalizing to different environments.

There exists many different LLMs, with different designers, architectures, training data, training strategies, number of parameters, reasoning mechanisms, etc. The first few LLM-based HAR approaches proposed have shown that the choice of the model has a big impact on HAR performance \cite{civitarese2025large, chen2024towards}. However, larger LLMs require significant computing power and induce high energy consumption. This prevents their deployment directly in home gateways or on other devices of the home (such as the smartphones of occupants), which is desirable to provide more guarantees on privacy.

In this work, we experimentally study the impact of LLM size on the performance of LLM-based HAR. In order to fairly compare the performance of different model sizes, we use the same family of models (Qwen3) and the same simple prompts for all models. These prompts may not necessarily lead to the best possible results on each dataset, but ensure that all models are provided with the same information and that even smaller models will understand the task described. We also study the potential of knowledge distillation approaches to fine-tune smaller models from reasoning examples generated by the largest models. Our goal is to show whether smaller language models (sometimes called SLMs), which could potentially be deployed locally in homes, can be improved to perform HAR as well as the largest models that require high computing power.

The main contributions of this study are:
\begin{enumerate}
    \item We report and discuss experimental results on HAR using a series of 6 LLMs from the Qwen3 family of models \cite{yang2025qwen3technicalreport}, ranging from 0.6B to 32B parameters. These results, on 2 state-of-the-art multi-subject HAR datasets, illustrate the large impact of model size on HAR performance.
    \item We experiment on the use of knowledge distillation to fine-tune Qwen3-0.6B and Qwen3-1.7B from reasoning examples produced by Qwen3-32B. We show in our experiments that this approach allows smaller models to perform almost as well as bigger models, even in cross-dataset scenarios, while using only a fraction of the computational cost of bigger models.
\end{enumerate}

\section{State of the Art}
\label{section:sota}

In this section, we review the current state of the art on LLM-based HAR, and knowledge distillation techniques for LLMs in general.

\subsection{LLM-Based HAR}

The research field of LLMs has been growing very rapidly in the last couples of years, with many different models being published: ChatGPT~\cite{achiam2023gpt}, Llama~\cite{touvron2023llama}, Qwen~\cite{yang2025qwen3technicalreport}, DeepSeek~\cite{guo2025deepseek}, etc. The use of LLMs for HAR is thus very recent, and few works have been published as of yet \cite{arrotta2025multi}. LLMs offer promising capabilities for generalization, in-context few-shot learning, and unsupervised HAR.

Takeda et al. \cite{10179111} first used LLMs to generate predictions of future sensor events based on ongoing activity and previous events. This first work showcased the potential of leveraging LLMs to infer information between sensors and activities. Gao et al. \cite{gao2024unsupervised} were the first to explicitly use LLM for HAR. They showed that a chain-of-thought \cite{wei2022chain} approach on windows of sensor events, integrating context information about the home, allowed an LLM to reach state-of-the-art HAR performance, despite being unsupervised unlike state-of-the-art models. Civitarese et al. \cite{civitarese2025large} proposed a more extensive experimental study of the capabilities of LLMs for zero-shot and few-shot HAR. This work highlights once again that LLMs can reach state-of-the-art performance without any training data, and that few-shot learning can improve performance. Chen et al. \cite{chen2024towards} used LLMs for multi-subject HAR, where situations can involve multiple users simultaneously in the home. This work showed that LLMs can correctly infer multi-subject activities from sensor events, and assign them to their corresponding users, in an unsupervised fashion.

\subsection{Knowledge Distillation for LLM}

Knowledge distillation offers the possibility of transferring the capabilities of one LLM (the teacher model) to another (the student model) \cite{xu2024surveyknowledgedistillationlarge}. One of the main applications of this approach consists in transferring the reasoning and planning capabilities of the most advanced models (which are often proprietary) to open-source models. Another application of knowledge distillation consists in transferring these capabilities from the most complex models to smaller models, in an attempt to reduce the computing requirements of a task while keeping the performance of large models. There exists many different distillation algorithms and techniques \cite{xu2024surveyknowledgedistillationlarge}. The most straightforward yet effective approaches is supervised fine-tuning: the student model's parameters are fine-tuned to align its predictions with the predictions of the teacher model.

As far as we know, knowledge distillation has not yet been experimented on in the context of HAR at home. In this work, we will thus aim at providing first results on the matter, focusing on the simple supervised fine-tuning approach to obtain baseline results.

\section{Method}

\subsection{Multi-user HAR with LLM}
\label{section:har}

We first seek to evaluate the impact of LLM size (i.e. the number of parameters of the model) on the task of HAR in general. Our goal is not necessarily to optimize a specific approach to reach optimal results on specific datasets. In fact, we prefer to use a straightforward approach for LLM-based HAR, with as few elements as possible that are specific to this approach. This way, we hope that the experimental results will reflect mostly on the general capabilities of LLMs for HAR, and will reflect as less as possible on the specific choices made in this study.

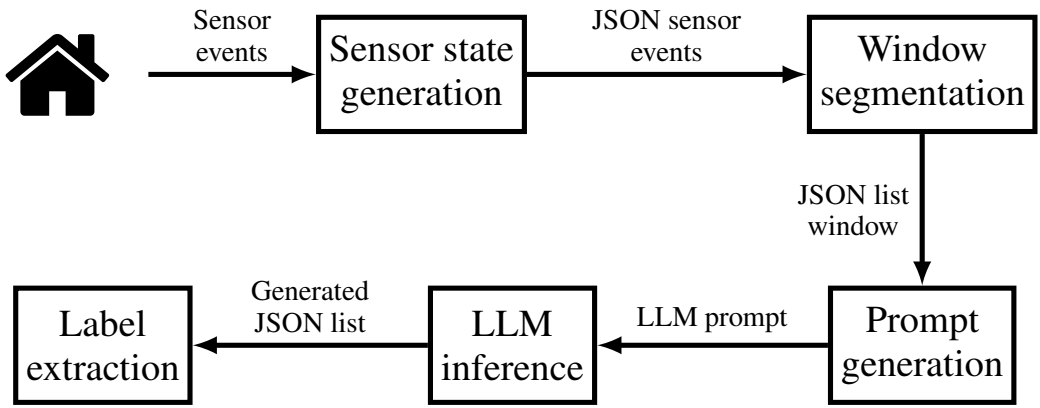
\begin{figure*}[t]
    \centering
    \input{plot_approach}
    \vspace{0.2cm}
    \caption{Architecture of the direct LLM-based HAR approach used in our experiments.}
    \label{fig:approach}
\end{figure*}

Following this goal, we implement a direct LLM-based HAR approach, which we illustrate in Figure \ref{fig:approach}. This approach only includes the main steps necessary to perform HAR with an LLM. These steps are common to many state-of-the-art approaches already mentioned such as \cite{chen2024towards, civitarese2025large}. We describe each step in the following subsections below.

All models used in our experiments use this same exact framework, with identical prompts, to ensure fair comparison. The prompt given to the model is intentionally simple and short, in the hopes of facilitating the understanding of the prompt for smaller models, to the detriment of performance on larger models.

\subsubsection{Sensor state generation}

For each raw sensor event collected in the home, we generate a corresponding JSON object containing the ID of the event (starting at 0, increasing), the time of the event (in \texttt{HH:mm:ss} format), and a textual description of the event. This textual description follows the following format: ``\texttt{<room\_name> <sensor\_name> <event>}'', where \texttt{<room\_name>} is the location of the sensor and where \texttt{<event>} depends on the type of the sensor event measured (e.g. ``\texttt{turned ON/OFF}'', ``\texttt{OPENED/CLOSED}'', etc.).

\subsubsection{Window segmentation}

In order to limit the context size given as input to the LLM, we split the resulting JSON list of events into non-overlapping windows of fixed-size. With small-sized windows, the LLM will infer the activity corresponding to each event with limited information about the other sensor events occurring before and after the current one. With large windows, the LLM can leverage a lot of information about previous and future events to infer activity, but might rely too much on distant context which is not relevant for the current sensor event.

\subsubsection{Prompt generation}

For each window (i.e. a sub-list of JSON events), a prompt is created with the following structure\footnote{The full prompt for MuRAL can be found at \url{https://pastebin.com/aqiwyDLf} and for Marble at \url{https://pastebin.com/3bnLcR2A}}:
\begin{itemize}
    \item a static description of the role of the LLM and the general multi-user HAR task,
    \item a description of the rooms of the home and the general appliances they contain (changes depending on the dataset),
    \item a static description of the input JSON structure and expected output for the HAR task,
    \item a list of all possible activity labels the model can predict (changes depending on the dataset),
    \item an optional list of a few rules to help infer activities in ambiguous situations (changes depending on the dataset),
    \item a static example of the structure of a given input and the structure of an expected output.
\end{itemize}

\subsubsection{LLM inference}

The model is given the previously generated prompt for each window, and generates a corresponding JSON list of activity label predictions for each sensor event in the window, as described in the prompt.

\subsubsection{Label extraction}

The generated JSON list of activity label predictions is parsed to extract the predicted activity for each sensor event, which can then be used to evaluate the performances of the model. In cases where the model generated a malformed output, or if the model did not generate a prediction for some sensor events of the window, then the prediction is considered incorrect for these events.

\subsection{Fine-Tuned HAR with LoRA}
\label{section:finetuning}

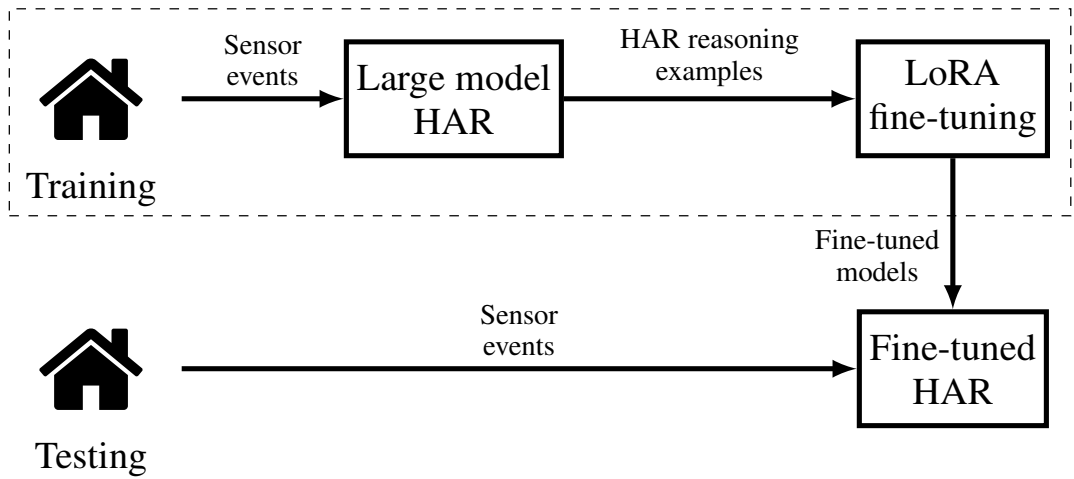
\begin{figure*}[t]
    \centering
    \input{plot_lora}
    \vspace{0.2cm}
    \caption{Architecture of the proposed knowledge distillation approach based on LoRA fine-tuning. Qwen3-32B is used as teacher, and Qwen3-0.6B / Qwen3-1.7B as students.}
    \label{fig:lora}
\end{figure*}

Our second goal in this work consists in studying the potential of knowledge distillation techniques for LLM-based HAR. More specifically, we choose to experiment on a direct supervised fine-tuning approach: a large model acts as teacher, generating HAR reasoning examples on a training set; a smaller model acts as a student, being fine-tuned to replicate these reasoning examples. The resulting fine-tuned smaller model can then be evaluated on HAR tasks like other models. The end goal of these experiments is to study whether this fine-tuning approach can help smaller models learn reasoning strategies employed by the larger, more accurate models. This in turn would significantly improve the performance of small models, while keeping a number of parameters (and thus computing requirements) orders of magnitude smaller than the larger models.

This approach is illustrated in Figure \ref{fig:lora}. We describe each step in the following subsections below.

\subsubsection{Large model HAR}

HAR is performed following the approach presented in Section \ref{section:har} and Figure \ref{fig:approach}, using the largest LLM available (in our case, Qwen3-32B), on training data. This training data is generally a subset of a larger dataset which also contains the separate evaluation data, but we also consider the case where the training data have been recorded in a completely different home environment (we call this scenario ``cross-dataset'' in the rest of the paper).

This HAR step leads to the generation of natural language reasoning by the LLM to infer activities for each window. Instead of only parsing the inferred activities to evaluate the model, we store the full reasoning texts generated by the model.

\subsubsection{LoRA fine-tuning}

The reasoning examples generated in the previous step are used to fine-tune a smaller LLM (in our case, Qwen3-0.6B and Qwen3-1.7B). We use a LoRA \cite{hu2022lora} fine-tuning approach following the support of state-of-the-art studies discussed in Section \ref{section:sota} such as \cite{ghosh2024closer}. LoRA’s base principle consists in freezing the weights $W_0$ of the original model, and adding trainable low-rank matrices $A$ and $B$, which change the forward pass of a neural unit $h = W_0x$ to $h = W_0x + BAx$. This way, the model can adjust its outputs and thus be fine-tuned by only training these matrices, which contain orders of magnitude less parameters than the base model. Typically in transformers, on which current LLMs are based, low-rank matrices are added only to the attention modules, and not to the MLP modules.

Smaller student models are thus fine-tuned in a supervised fashion using LoRA. The student model is trained to replicate the reasoning outputs generated by the large teacher model for all windows of training data.

In this paper, we do not perform any filtering on the examples generated by the large model. All reasoning examples for all windows of the training dataset are used for the fine-tuning step. Future work could study the impact of selecting high quality reasoning examples, for example excluding reasonings that lead to incorrect classification of activities.

\subsubsection{Fine-tuned HAR}

The resulting fine-tuned model is then evaluated as described in Section \ref{section:har} on the evaluation data. This model is slightly larger than the original small student model, due to the addition of low-rank matrices for LoRA fine-tuning. In our experiments, the increase in number of parameters is $3.25\%$ for Qwen3-0.6B and  $1.99\%$ for Qwen3-1.7B. The number of parameters of the fine-tuned models are thus comparable to the original models, and much smaller than the large teacher model.

\section{Experimental Results}

\subsection{Datasets}

To evaluate the impact of model size and knowledge distillation on LLM-based HAR, we use two state-of-the-art multi-subject HAR datasets: Marble \cite{arrotta2021marble} and MuRAL\cite{chen2025muralmultiresidentambientsensor}. Both datasets contain scenarios of activities of daily-living in instrumented homes, with up to 4 subjects present in the environment simultaneously. Marble contains data from both environmental sensors and smart-watches, while MuRAL contains only environmental data. 

Multi-subject scenarios recorded in these datasets lead to more complex situations than single-subject datasets: the number of inhabitants is not known by the model a priori, and collected data from environmental sensors are difficult to assign to a specific subject and activity. Therefore, larger models should be able to outshine smaller models on the task of inferring activities from sensor data. Moreover, multi-subject situations are more representative of real-world scenarios of family life at home, and it is thus more valuable to report the impact of model size for such cases.

In this paper, we only focus on the correctness of predicted activities, and not on the association of the predicted activity to the correct resident.

\subsection{Frameworks and Hardware}

All LLM inferences were run using vLLM \cite{kwon2023efficient}. The LoRA fine-tuning training step was performed using Unsloth \cite{unsloth}. All experiments were run using NVIDIA A100 GPUs.

\subsection{Experimental protocol}

\subsubsection{Data separation}

We divided both datasets with the same strategy: all reported experimental results were evaluated on roughly the last third of each dataset, while the first two thirds were reserved as training set for the fine-tuning experiments.

\begin{itemize}
    \item In Marble, each scenario contains between 2 to 4 recording sessions. For 2-session scenarios, the first one was kept for training, and the last one for testing. For 3 and 4-session scenarios, the first two were kept for training, and the last one or two for testing. This way, each scenario is represented fairly in both training and testing sets.
    \item In MuRAL, there are 21 independent recording sessions. The first 15 sessions were kept for training, while the last 6 sessions were used for testing.
\end{itemize}

For both datasets, we divided each session in sliding windows (with no overlap) containing 10 consecutive sensor events. For each window, the model infers the activity for each of the 10 sensor events. The final reported performance is the F1-Score computed over all sensor event predictions, averaged for each session in the test set.

\subsubsection{Fine-tuning} We used the reasoning examples generated by Qwen3-32B on all windows of training set sessions (126 windows in total for Marble, 544 for MuRAL), with no filtering. This way, no bias is introduced through selection criteria for choosing reasoning examples to give to the student model. Following preliminary evaluations on the training data only, We fine-tune Qwen3-0.6B and Qwen3-1.7B with a limit of 500 steps (which corresponds to 32 epochs of training for Marble, and 8 for MuRAL), and an initial learning rate of $2e^{-4}$.




\subsection{Comparison of LLMs with different sizes}

\begin{table}[t]
\centering
\caption{F1-Scores (in \%) of different sizes of Qwen3.}
\vspace{0.2cm}
\begin{tabular}{|l|c|c|c|}
\hline
\multicolumn{1}{|c|}{Model} & Marble & MuRAL \\ 
\hline
Qwen3-0.6B & 16.52 & 10.81 \\
Qwen3-1.7B & 26.73 & 12.42 \\
Qwen3-4B & 32.57 & 35.67 \\
Qwen3-8B & 36.32 & 46.29 \\
Qwen3-14B & 36.74 & 48.82 \\
Qwen3-32B & 37.61 & 53.70 \\
\hline
\end{tabular}
\label{table:size_comparison}
\end{table}

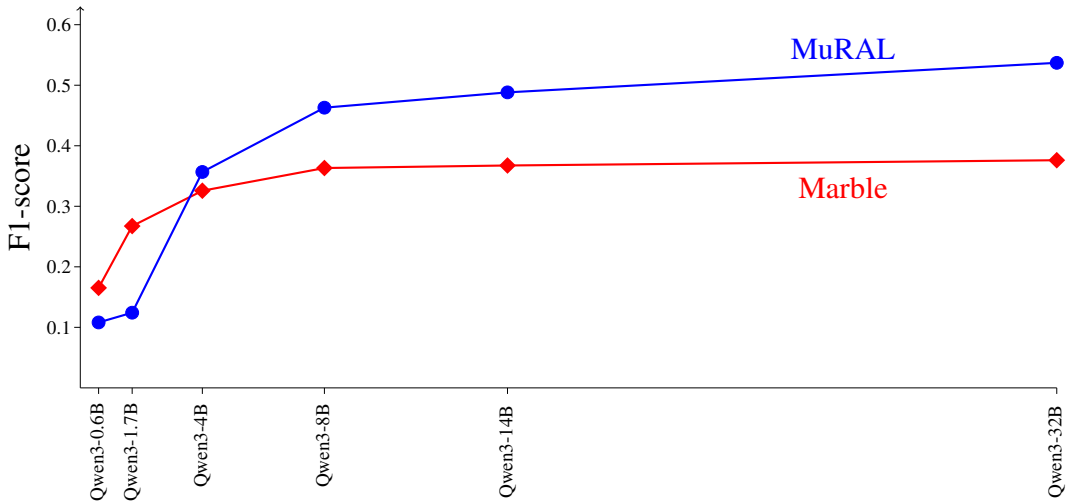
\begin{figure*}[t]
    \centering
    \input{plot_comparison}
    \vspace{0.2cm}
    \caption{Performance of different sizes of Qwen3. The x-axis is to scale with the number of parameters of each model.}
    \label{fig:size_comparison}
\end{figure*}

We report in Table \ref{table:size_comparison} the performance of 6 different Qwen3 models, from 0.6B to 32B parameters, on the Marble and MuRAL datasets. We see a clear correlation between the number of parameters of models and the performance for multi-subject HAR, on both datasets. Qwen3-0.6B performs as low as 10.81\% on MuRAL, while Qwen3-32B reaches 53.70\% F1-Score.

As illustrated in Figure \ref{fig:size_comparison}, the evolution of performances with respect to the number of parameters of each model is clearly not linear. For Marble, performance grows steadily until it reaches a plateau with Qwen3-8B. Qwen3-14B and Qwen3-32B still improve performances, but the gains are much smaller considering the increase in the number of parameters. For MuRAL, both Qwen3-0.6B and Qwen3-1.7B exhibit similar low performance, which then increases significantly with Qwen3-4B. A similar plateau is reached with Qwen3-8B, although the increase in performance with Qwen3-14B and Qwen3-32B are more significant than on Marble.

On both datasets, we thus observe the strong impact of model size on multi-subject HAR performance, where performance increases every time a larger model is used. However, we see that these performances are not linearly distributed. Smaller models (Qwen3-0.6B, Qwen3-1.7B) perform very poorly, especially on MuRAL. Qualitative studies of the reasoning performed by these models shows that they sometimes struggle to understand the tasks and directions given in the prompt (despite our choice of using simple, short prompts), and that the reasoning hypotheses they use often do not follow any kind of logical sense, which leads to inaccurate activity recognition. On the other hand, larger models (Qwen3-14B, Qwen3-32B) clearly perform the best and showcase logical reasoning, but this only marginally improves the performance compared to Qwen3-8B, despite much greater computing costs.

\subsection{Performances After Fine-Tuning}
\label{section:exp_finetuning}

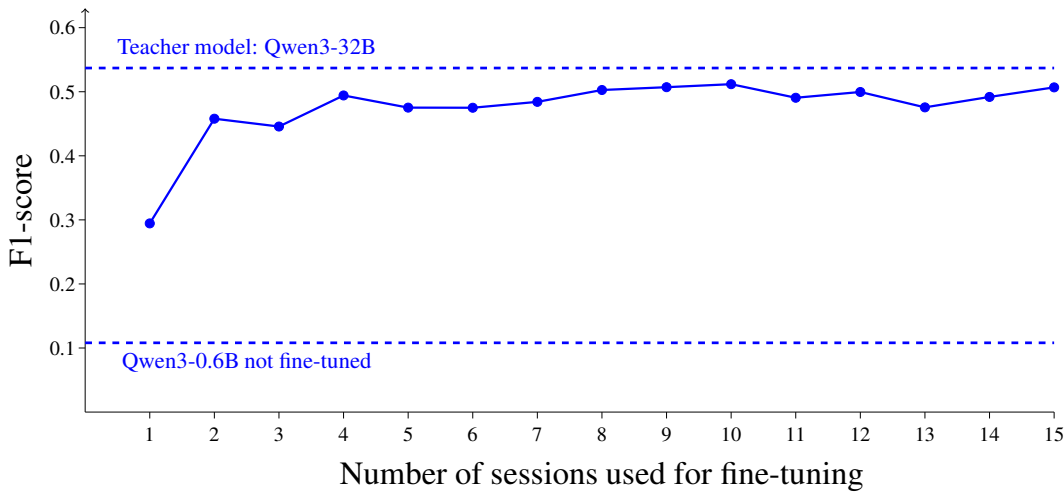
\begin{figure*}[t]
    \centering
    \input{plot_finetune_sessions}
    \vspace{0.2cm}
    \caption{Fine-tuned Qwen3-0.6B performance on MuRAL depending on the number of MuRAL sessions used for fine-tuning.}
    \label{fig:finetuning}
\end{figure*}

\begin{table*}[t]
\centering
\caption{F1-Scores (in \%) of fine-tuned models, with increase ($\uparrow$) or decrease ($\downarrow$) in performance compared to non-fine-tuned models (see Table \ref{table:size_comparison}).}
\begin{tabular}{|l|l|l|l|}
\hline
\multicolumn{1}{|c|}{Model} & \multicolumn{1}{|c|}{Marble} & \multicolumn{1}{|c|}{MuRAL} \\ 
\hline
Qwen3-0.6B fine-tuned on Marble & 32.38 ($\uparrow$ 15.86) & 9.66 ($\downarrow$ 1.15) \\
Qwen3-0.6B fine-tuned on MuRAL & 32.05 ($\uparrow$ 15.53) & 50.68 ($\uparrow$ 39.87) \\
Qwen3-1.7B fine-tuned on Marble & 33.55 ($\uparrow$ 6.82) & 16.25 ($\uparrow$ 3.83) \\
Qwen3-1.7B fine-tuned on MuRAL & 34.27 ($\uparrow$ 7.54) & 52.67 ($\uparrow$ 41.28) \\
\hline
\end{tabular}
\label{table:finetuned}
\end{table*}

We report in Table \ref{table:finetuned} the performances of Qwen3-0.6B and Qwen3-1.7B, following a step of fine-tuning as described in Section \ref{section:finetuning}. Both models were fine-tuned using reasoning examples generated by Qwen3-32B, on the training data sets of Marble and MuRAL. We can see a major improvement in performances following this fine-tuning step, compared to the non-fine-tuned results of Qwen3-0.6B and Qwen3-1.7B reported in Table \ref{table:size_comparison}. On Marble, Qwen3-0.6B and Qwen3-1.7B reach F1-Scores of 32.38\% and 33.55\% respectively, which are slightly lower than the teacher model Qwen3-32B (37.61\%), but much higher than the non-fine-tuned results (16.52\% and 26.73\% respectively). Similarly for MuRAL, the performances of the fine-tuned Qwen3-0.6B and Qwen3-1.7B (50.68\% and 52.67\% respectively) are close to the performance of Qwen3-32B (53.70\%), and a huge improvement on the non-fine-tuned results (10.81\% and 12.42\% respectively).

\subsection{Influence of Training Set Size on Fine-Tuned Models}

We report in Figure \ref{fig:finetuning} the performance of fine-tuned Qwen3-0.6B depending on the number of sessions used from the training set, on the MuRAL dataset (e.g. for 3 sessions, the teacher Qwen3-32B generated reasoning examples on the first 3 sessions of MuRAL only, instead of the full 15 sessions). 

We see that, even with only one session of reasoning examples, the increase in performance compared to the non-fine-tuned model is already very significant (29.45\% compared to 10.81\%). A performance plateau is quickly reached at 4 sessions (49.42\%), after which performance stays relatively stable regardless of the number of additional sessions used. The best performance is actually reached at 10 sessions used (51.18\%), which is slightly higher than when using all 15 sessions (50.68\%). Regardless of the number of sessions used, the performance of the fine-tuned model is always lower than the teacher model Qwen3-32B.

Overall, we observe in this experiment that the amount of data required to fine-tune a smaller model from reasoning examples of a larger model is not necessarily large. Only a fraction of the MuRAL dataset suffices to reach a plateau of performance slightly below the performance of the teacher model. These results suggest that, for most activity instances, the number of reasoning patterns employed by the teacher model to infer activities is quite limited, and the student model can thus learn these reasoning patterns with limited examples.

\subsection{Cross-Dataset Fine-Tuning}

We also report in Table \ref{table:finetuned} the performances of fine-tuned models in cross-dataset scenarios: testing, on the MuRAL dataset, a model that was fine-tuned using reasoning examples generated on the training set of the Marble dataset (and vice-versa). 

We observe two different behaviors. Qwen3-0.6B and Qwen3-1.7B, fine-tuned on the MuRAL dataset, perform as well on the Marble dataset (32.05\% and 34.27\%) as their counterparts fine-tuned directly on the Marble dataset (32.38\% and 33.55\%). This supports the observations discussed in Section \ref{section:exp_finetuning} that this fine-tuning step transfers the reasoning strategies to infer activities, more so than transferring overfitting information about the specific training dataset used. Overall, these results suggest that such a fine-tuning approach is promising for generalization, by leveraging reasoning examples generated on sufficiently rich datasets to fine-tune models for any other environment of application.

On the other hand, Qwen3-0.6B and Qwen3-1.7B fine-tuned with the Marble dataset do not perform well on the MuRAL dataset (9.66\% and 16.25\% respectively). Performance is slightly worse than the non-fine-tuned model for Qwen3-0.6B (9.66\% vs 10.81\%), and slightly better than the non-fine-tuned model for Qwen3-1.7B (16.25\% vs 12.42\%). This apparent difference with previous results could be explained from the complexity of the dataset used for fine-tuning: MuRAL contains approximately 5 times as many data windows as Marble, and all situations are multi-user, whereas many situations in Marble are single-user. Moreover, a significant portion of data in Marble comes from wearable sensors, which are very informative when recognizing single-user activities, but reasoning on such data cannot be leveraged for MuRAL, where only environmental sensors are used. Conversely, reasoning on complex environmental data learned on MuRAL can be leveraged on Mable, which also contains simpler environmental data.

\subsection{Reasoning comparisons}

\begin{table*}[t]
    \centering
    \caption{Reasoning examples of Qwen3-0.6B, Qwen3-32B, and Qwen3-0.6B fine-tuned on MuRAL, on the same test event from the MuRAL dataset (only a short extract of the reasoning is included).}
    \resizebox{\linewidth}{!}{
    \begin{tabularx}{\linewidth}{|*{3}{X|}}
    \hline
    \multicolumn{3}{|l|}{\textbf{Event}: \scalebox{.9}{\texttt{\{time: "19:53:04", event: "kitchen induction stove wattmeter turned OFF"\}}}}\\
    \multicolumn{3}{|l|}{\textbf{Ground truth label}: \texttt{"19. preparing dinner"}}\\
    \hline
    \multirow{2}{*}{\begin{tabular}{c}Qwen3-0.6B\end{tabular}} & \multirow{2}{*}{\begin{tabular}{c}Qwen3-32B\end{tabular}} & \multirow{2}{*}{\begin{tabular}{c}Qwen3-0.6B\\fine-tuned on MuRAL\end{tabular}} \\
    & & \\
    \hline
    […] The activity list includes ``personal washing'' as 25. So maybe the subject is someone in the kitchen, and activity is 25. […] 
    & […] Turning it off might indicate that the person was cooking, maybe preparing dinner. So activity 19. […] 
    & […] The induction stove is part of the kitchen, and turning it off could indicate that someone has finished cooking. […] Let's tentatively assign this to preparing dinner (19) […]
    \\
    \hline
    \end{tabularx}
    }
    \label{table:reasoning}
\end{table*}

We report in Table \ref{table:reasoning} an example of reasoning of different models on the same test event from the MurAL dataset. Only short extracts of each reasoning are included for the sake of brevity. This example event involves a kitchen induction stove being turned off, at around 8 p.m., and the ground truth activity label of this event is ``19. preparing dinner''. We see that Qwen3-32B correctly infers the right activity, noticing that the kitchen stove should indicate cooking-related activities. On the other hand, we see that Qwen3-0.6B incorrectly infers the activity as ``25. personal washing'', despite expressing that the subject might be in the kitchen. Multiple other examples of such reasoning can be found in entire evaluation results, where Qwen3-0.6B correctly identifies important contextual cues (such as the event being related to the kitchen), but infers activity that are not at all related to the previously identified cues. In short, Qwen3-0.6B does not seem to often follow coherent reasoning patterns, as opposed to larger models like Qwen3-32B.

Qwen3-0.6B fine-tuned on MuRAL, on the other hand, exhibits completely different reasoning patterns, which are much closer to those of the teacher Qwen3-32B model. Through these new reasoning strategies, it is not surprising that the fine-tuned model greatly improves its performance.

\subsection{Missed events}

As mentioned in Section \ref{section:har}, if the LLM did not infer an activity for a certain event, then its prediction for this event was considered incorrect. This can be caused by 3 main issues originating from the model itself: 
\begin{itemize}
    \item the model can forget about some events of the window throughout its reasoning, leading to missing events in its answer,
    \item the model can infer an activity for all events of the window, but does not report all of them in its answer,
    \item the model can generate a malformed answer that does not follow the proper JSON format described in the prompt.
\end{itemize}

\begin{table*}[t]
\centering
\caption{Percentages of events not classified by each model, either because of incomplete reasoning, incomplete answers, or malformed answers.}
\begin{tabular}{|c|c|c|c|}
\hline
\multicolumn{1}{|c|}{Model} & Marble & MuRAL \\ 
\hline
Qwen3-0.6B & 13.52 ± 16.85 & 16.66 ± 8.84 \\
Qwen3-1.7B & 2.26 ± 3.91 & 2.49 ± 0.44 \\
Qwen3-4B & 7.07 ± 14.94 & 0.19 ± 0.10\\
Qwen3-8B & 2.25 ± 8.25 & 0.13 ± 0.14 \\
Qwen3-14B & 4.60 ± 12.44 & 0.13 ± 0.14 \\
Qwen3-32B & 0.0 ± 0.0 & 0.13 ± 0.14\\
\hline
Qwen3-0.6B fine-tuned on Marble & 0.63 ± 1.73 & 7.21 ± 3.52 \\
Qwen3-0.6B fine-tuned on MuRAL & 0.77 ± 4.28 & 1.83 ± 1.99 \\
Qwen3-1.7B fine-tuned on Marble & 0.09 ± 0.50 & 2.03 ± 1.49 \\
Qwen3-1.7B fine-tuned on MuRAL & 0.0 ± 0.0 & 0.58 ± 0.95 \\
\hline
\end{tabular}
\label{table:missed}
\end{table*}

We report in Table \ref{table:missed} the percentages of events for which each model did not answer with an activity prediction. We see that, for non-fine-tuned models, the number of missed events and standard deviation generally decreases with the size of the LLM (results are a bit more erratic for Marble). This supports the expected idea that larger models tend to follow instructions more faithfully (reducing the number of malformed answers), and hold a more steady account of the list of events given in the current window. For fine-tuned models, we see that the number of missed events decreases drastically compared to the original models, once again showcasing that this fine-tuning approach greatly improves the reasoning processes of smaller models.

\subsection{Discussion}

Our first experimental results on Marble and MuRAL suggest that fine-tuning is a promising avenue for reducing the computing costs of LLM-based HAR. Fine-tuned models can reach performances that are very close to the original teacher model, with only a few sessions of data. 

However, we also see that the choice of training data used has a great influence on the performance of the fine-tuned models: reasoning strategies produced by the teacher model should be as generalizable as possible to all environments, and not rely on specific situations or data sources. Moreover, the performance of the teacher model is also crucial: in all our experiments, fine-tuned models could approach the performance of the teacher model, but never reach or surpass it. In a sense, the performance of the teacher model appears to be the upper bound of attainable performance for fine-tuned models.

\section{Conclusion}

In this paper, we presented an experimental study of the impact of LLM size on the performance of LLM-based HAR. We have shown, on two state-of-the-art datasets, that large models perform better at HAR than small models of the Qwen3 family of models, as expected. However, performance improvement is not linear with respect to the number of parameters of models, and we observe diminishing returns in performance as models grow in size (and in required computing power).

We also experimented on the use of knowledge distillation techniques to fine-tune smaller models from reasoning examples generated by larger models. We have shown on the same two datasets that the smallest Qwen3-0.6B model can reach HAR performance close to the teacher Qwen3-32B model following this fine-tuning step, even in cross-dataset scenarios. Only a fraction of the available training data is required to reach high performance with the fine-tuned model, suggesting that the student model correctly learns the general reasoning strategies of larger models, instead of overfitted details.

These experiments suggest that future HAR systems could rely on small language models, fine-tuned with a limited set of HAR reasoning examples generated with larger models on representative data. Such an approach would allow to combine the accuracy, context-awareness, and generalization capabilities of large models for HAR, with the resource frugality of smaller models. Future work should focus on extending these experiments with more datasets and models, striving to reduce further the size of language models used for HAR. As it stands, Qwen3-0.6B is still too large to be appropriately deployed on home gateways or other candidate devices for local deployment.

\bibliographystyle{splncs04}
\bibliography{refs}

\end{document}

%% file: plot_approach.tex
\resizebox{0.8\linewidth}{!}{

\begin{tikzpicture}[font=\large]

\node[scale=3] at (-0.5,0) {\faHome};
\node[draw, ultra thick, align=center, minimum height=36pt] at (3.5,0) (1) {Sensor state\\generation};
\node[draw, ultra thick, align=center, minimum height=36pt] at (9,0) (2) {Window\\segmentation};
\node[draw, ultra thick, align=center, minimum height=36pt] at (9,-3) (3) {Prompt\\generation};
\node[draw, ultra thick, align=center, minimum height=36pt] at (4.5,-3) (4) {LLM\\inference};
\node[draw, ultra thick, align=center, minimum height=36pt] at (0,-3) (5) {Label\\extraction};

\draw[-latex, ultra thick] (0.5,0) -- node [above, align=center, font=\small] {Sensor\\events} (1.west);
\draw[-latex, ultra thick] (1.east) -- node [above, align=center, font=\small] {JSON sensor\\events} (2.west);
\draw[-latex, ultra thick] (2.south) -- node [left, align=center, font=\small] {JSON list\\window} (3.north);
\draw[-latex, ultra thick] (3.west) -- node [above, align=center, font=\small] {LLM prompt} (4.east);
\draw[-latex, ultra thick] (4.west) -- node [above, align=center, font=\small] {Generated\\JSON list} (5.east);

\end{tikzpicture}
}

%% file: plot_lora.tex
\resizebox{0.8\linewidth}{!}{

\begin{tikzpicture}[font=\large]

\node[scale=3] at (-0.5,0) {\faHome};
\node at (-0.5,-1) {Training};
\node[draw, ultra thick, align=center, minimum height=36pt] at (3.5,0) (1) {Large model\\HAR};
\node[draw, ultra thick, align=center, minimum height=36pt] at (9,0) (2) {LoRA\\fine-tuning};

\draw[dashed] (-1.4,1) rectangle ++(11.7,-2.3);

\node[scale=3] at (-0.5,-3) {\faHome};
\node at (-0.5,-4) {Testing};
\node[draw, ultra thick, align=center, minimum height=36pt] at (9,-3) (3) {Fine-tuned\\HAR};

\draw[-latex, ultra thick] (0.5,0) -- node [above, align=center, font=\small] {Sensor\\events} (1.west);
\draw[-latex, ultra thick] (1.east) -- node [above, align=center, font=\small] {HAR reasoning\\examples} (2.west);
\draw[-latex, ultra thick] (2.south) -- node [left, align=center, font=\small] {\\\\Fine-tuned\\models} (3.north);

\draw[-latex, ultra thick] (0.5,-3) -- node [above, align=center, font=\small] {Sensor\\events} (3.west);

\end{tikzpicture}
}

%% file: plot_comparison.tex
\resizebox{0.8\linewidth}{!}{
\begin{tikzpicture}[font=\Large]

  \def\yScale{10};
  \def\xScale{0.5};

  \draw[-] (0,0) -- (32*\xScale,0);
  \draw[->] (0,0) -- (0,{0.63*\yScale});

  \node[rotate=90] at (-1, {0.315*\yScale}) {F1-score};

    \draw ({0.6*\xScale},0) -- ({0.6*\xScale},-0.1); 
    \node [rotate=90, xshift=-0.7cm] at ({0.6*\xScale}, -0.35) {\small Qwen3-0.6B}; 

    \draw ({1.7*\xScale},0) -- ({1.7*\xScale},-0.1); 
    \node [rotate=90, xshift=-0.7cm] at ({1.7*\xScale}, -0.35) {\small Qwen3-1.7B}; 

    \draw ({4*\xScale},0) -- ({4*\xScale},-0.1); 
    \node [rotate=90, xshift=-0.6cm] at ({4*\xScale}, -0.35) {\small Qwen3-4B}; 

    \draw ({8*\xScale},0) -- ({8*\xScale},-0.1); 
    \node [rotate=90, xshift=-0.6cm] at ({8*\xScale}, -0.35) {\small Qwen3-8B}; 

    \draw ({14*\xScale},0) -- ({14*\xScale},-0.1); 
    \node [rotate=90, xshift=-0.7cm] at ({14*\xScale}, -0.35) {\small Qwen3-14B}; 

    \draw ({32*\xScale},0) -- ({32*\xScale},-0.1); 
    \node [rotate=90, xshift=-0.7cm] at ({32*\xScale}, -0.35) {\small Qwen3-32B}; 

  \foreach \y in {0.1,0.2,0.3,0.4,0.5,0.6} {
    \draw (0, {\y*\yScale}) -- (-0.1, {\y*\yScale}); 
    \node at (-0.35, {\y*\yScale}) {\small \y}; 
  }

  \foreach \x/\y in {
    {0.6*\xScale}/{0.1652*\yScale},
    {1.7*\xScale}/{0.2673*\yScale},
    {4*\xScale}/{0.3257*\yScale},
    {8*\xScale}/{0.3632*\yScale},
    {14*\xScale}/{0.3674*\yScale},
    {32*\xScale}/{0.3761*\yScale}
  } {
    \node[diamond, fill=red, inner sep=0, minimum size=8pt] at (\x,\y) {};
  }
  \draw[line width=1pt, red] ({0.6*\xScale},{0.1652*\yScale}) -- ({1.7*\xScale},{0.2673*\yScale}) -- ({4*\xScale},{0.3257*\yScale}) -- ({8*\xScale},{0.3632*\yScale}) -- ({14*\xScale},{0.3674*\yScale}) -- ({32*\xScale},{0.3761*\yScale});

  \foreach \x/\y in {
    {0.6*\xScale}/{0.1081*\yScale},
    {1.7*\xScale}/{0.1242*\yScale},
    {4*\xScale}/{0.3567*\yScale},
    {8*\xScale}/{0.4629*\yScale},
    {14*\xScale}/{0.4882*\yScale},
    {32*\xScale}/{0.5370*\yScale}
  } {
    \filldraw[blue] (\x,\y) circle (3pt);
  }
  \draw[line width=1pt, blue] ({0.6*\xScale},{0.1081*\yScale}) -- ({1.7*\xScale},{0.1242*\yScale}) -- ({4*\xScale},{0.3567*\yScale}) -- ({8*\xScale},{0.4629*\yScale}) -- ({14*\xScale},{0.4882*\yScale}) -- ({32*\xScale},{0.5370*\yScale});

    \node [blue] at ({25*\xScale}, {0.56*\yScale}) {\Large MuRAL}; 
    \node [red] at ({25*\xScale}, {0.33*\yScale}) {\Large Marble}; 

\end{tikzpicture}
}

%% file: plot_finetune_sessions.tex
\resizebox{0.8\linewidth}{!}{
\begin{tikzpicture}[font=\Large]

  \def\yScale{10};

  \draw[-] (0,0) -- (15,0);
  \draw[->] (0,0) -- (0,{0.63*\yScale});

  \node at (8, -1) {Number of sessions used for fine-tuning};
  \node[rotate=90] at (-1, {0.315*\yScale}) {F1-score};

  \foreach \x in {1,2,3,4,5,6,7,8,9,10,11,12,13,14,15} {
    \draw (\x,0) -- (\x,-0.1); 
    \node at (\x, -0.35) {\small \x}; 
  }

  \foreach \y in {0.1,0.2,0.3,0.4,0.5,0.6} {
    \draw (0, {\y*\yScale}) -- (-0.1, {\y*\yScale}); 
    \node at (-0.35, {\y*\yScale}) {\small \y}; 
  }

  \foreach \x/\y in {
    1/{0.2945*\yScale},
    2/{0.4579*\yScale},
    3/{0.4457*\yScale},
    4/{0.4942*\yScale},
    5/{0.4752*\yScale},
    6/{0.4750*\yScale},
    7/{0.4842*\yScale},
    8/{0.5027*\yScale},
    9/{0.5071*\yScale},
    10/{0.5118*\yScale},
    11/{0.4906*\yScale},
    12/{0.4996*\yScale},
    13/{0.4756*\yScale},
    14/{0.4919*\yScale},
    15/{0.5068*\yScale}
  } {
    \filldraw[blue] (\x,\y) circle (2pt);
  }
  \draw[line width=1pt, blue] (1,{0.2945*\yScale}) -- (2,{0.4579*\yScale}) -- (3,{0.4457*\yScale}) -- (4,{0.4942*\yScale}) -- (5,{0.4752*\yScale}) -- (6,{0.4750*\yScale}) -- (7,{0.4842*\yScale}) -- (8,{0.5027*\yScale}) -- (9,{0.5071*\yScale}) -- (10,{0.5118*\yScale}) -- (11,{0.4906*\yScale}) -- (12,{0.4996*\yScale}) -- (13,{0.4756*\yScale}) -- (14,{0.4919*\yScale}) -- (15,{0.5068*\yScale});

  \draw[line width=1.2pt, dashed, blue] (0, {0.5370*\yScale}) -- (15, {0.5370*\yScale});
  \node at (2.5, {0.5370*\yScale + 0.3}) {\textcolor{blue}{\normalsize Teacher model: Qwen3-32B}};

  \draw[line width=1.2pt, dashed, blue] (0, {0.1081*\yScale}) -- (15, {0.1081*\yScale});
  \node at (2.5, {0.1081*\yScale - 0.3}) {\textcolor{blue}{\normalsize Qwen3-0.6B not fine-tuned}};


\end{tikzpicture}
}